# EVIDENTIAL REASONING IN IMAGE UNDERSTANDING

Mingchuan Zhang and Su-shing Chen
Department of Computer Science
University of North Carolina
Charlotte, NC 28223

## ABSTRACT

In this paper, we present some results of evidential reasoning in understanding multispectral images of remote sensing systems. The Dempster-Shafer approach of combination of evidences is pursued to yield contextual classification results, which are compared with previous results of the Bayesian context free classification, contextual classifications of dynamic programming and stochastic relaxation approaches.

## 1. INTRODUCTION

There has been a significant amount of research of evidential reasoning in symbolic and linguistic information processing. However, evidential reasoning in spatial information processing is only emerging to be a subject of interest recently. Spatial (2-D and 3-D) information processing is important to medical, military, engineering and industrial automation applications. In addition to the usual uncertainty in AI, uncertainty that arises from spatial information requires somewhat different consideration. In this paper, we shall study uncertainty in image understanding of multispectral images in remote sensing systems. Some experimental results of alternate approaches - the Bayesian context free classification, contextual classification using a dynamic programming approach, contextual classification using the stochastic relaxation approach, and the contextual classification using the Dempster-Shafer approach [1]-[5]. It turns out that the percentage of multispectral image classification accuracy is increasing in the above order.

There are two types of context information in real world images. One is local or pixel-based context information; the other is global or object-based context information. Most of the existing contextual classification methods have been developed using local context information on small neighborhoods [1]-[5]. One of the most difficult problems in remote sensing as yet unsolved is how to deal with "mixed pixels" effectively. In order to solve this problem and to meet increasing demand of classification accuracy, further research should focus on both local and global context information. General context information processing is potentially a powerful tool in spatial reasoning, for it is concerned with all kinds of information - local or global; certain or uncertain; complete or incomplete. Such a process may also incorporate relatively high-level intelligence in decision-making operations. In this paper, a new contextual reasoning method using Dempster-Shafer theory for multispectral image classification is proposed.

## 2. UNCERTAIN AND INCOMPLETE KNOWLEDGE IN MSS

A multispectral scanner system (MSS) provides spectral data in quantitative format over a broad range of wavelengths. Spatial features, such as size, shape, texture or linear feature, are extracted from pixel data which are the lowest level of image data. Classification methods are used to analyse these quantitative feature data. Data sets of this kind are often uncertain and incomplete, both in evidence and in world knowledge. For illustration, let us consider information of multispectral terrain reflectance. It is only a single parameter that is useful as an indicator of terrain classes. If the terrain within a pixel of the multispectral image is composed of a single feature, such as deep clear water, then the reflectance can be correlated in a high degree of confidence with a particular parameter of interest. However, the radiance received from the ground in a pixel of multispectral image is usually originated from a combination of soil, rock, vegetation, water, and man-made features within the pixel. Thus a pixel encompasses a variety of terrain features and the received radiance is the integration of the reflectance of all features. In this case, the Bayesian probabilistic model of random fields has some limitation and is not able to capture the full information due to incomplete evidences.

## 3. EVIDENTIAL REASONING

As mentioned in the previous section, multispectral scanner system probing an environment deals often with incomplete and uncertain information. In this paper, we shall use the Dempster-Shafer theory to represent uncertainty and to use the combination rule to reduce uncertainty and resolve contradictions.

Spatial information and knowledge are represented by propositions. These propositions may range from simple ones, such as "A certain region belongs to a particular terrain class", to high-level decision making ones, such as "Robot X should perform a particular task". For each proposition P, its belief is represented by an evidential interval [Spt(P), Pls(P)], where Spt(P) is the degree to which the evidence supports P and Pls(P) is the degree to which the evidence fails to refute P (the degree to which it remains plausible). A frame of discernment $\Theta$ is a set of propositions of mutually exclusive possibilities in a specific domain of a spatial reasoning system. The belief function Bel over $\Theta$ is defined by a probability distribution function m which is called the mass distribution in [6].

For the subsystem of contextual classification, regions of multispectral images are classified into a set of possible categories. First, a collection of spatial features are extracted from the scene. For each feature, there is a set of propositions to which it can directly contribute beliefs. Thus, a mass distribution m(f) is associated with each feature f over the frame $\Theta$ of discernment which is determined from ground truth or preclassified results. Furthermore, simple belief function is used for each feature f. That is, the belief function Bel(f) is the same as the mass distribution m(f) for each feature f.

If each extracted feature is considered as a piece of evidence, evidences of several features can be combined to be an accumulated evidence. The Dempster's combination rule is used to form orthogonal sum of several belief functions. Since this rule is associative and commutative, features can be combined in any order.

In our problem domain, bodies of evidence may point to different subsets of $\Theta$. This situation is called heterogeneous evidence in [6]. In the case of $A \cap B$ is not empty, the combination of two simple support functions $S_1$ and $S_2$, focused on A and B respectively, is carried out as follows. If $S_1(A) = s_1$ and $S_2(B) = s_2$, the Dempster's combination rule implies that $m(A \cap B) = s_1 s_2$, $m(A) = s_1(1-s_2)$, $m(B) = s_2(1-s_1)$, and $m(\Theta) = (1-s_1)(1-s_2)$. Recursively, this combination rule is extended to any number of features. In the case of $A \cap B$ is empty, the situation is called conflicting evidence in [6]. We refer to [6] for details of the rule.

The procedure of this evidential approach to contextual classification is as follows:
1. Partition ground truth or preclassification results into multiple bands. Each band corresponds to one labeling of the preclassification process.
2. Use the maximal connected component operation to label maximal connected components of each image band.
3. Extract all features and form a feature vector of each maximal connected component of an image band.

4. Select some regions as ground truth data. Determine mass distributions of all features of the ground truth data set for each image band.
5. For each region in an image band, generate hypothesis of classification.
6. Determine simple belief functions of all features of the given region.
7. Use the combination rule to compute the belief function of multi-features of the given region.
8. If the hypothesis is rejected, the region is merged to a neighboring region. The new region is tested. If the hypothesis is accepted, the remaining regions are tested.

## 4. SPATIAL FEATURES

The following spatial features are used in our classification system.

(1) Region Size. In MSS images, one pixel corresponds to 57 by 79 $M^2$ ground area. For instance, an image of one-crop farming fields (Figure 1) should have a finite number of fairly homogeneous regions. In view of the classification results of the Bayesian context free classification method and contextual classification methods [1], [2], there are misclassified isolated pixels or small regions, because they are assigned to different classes from neighboring homogeneous regions. This evidential reasoning scheme will enable us to verify the hypothesis of these misclassified pixels or regions. The mass distribution associated to the size feature of each image band is determined as follows. Using a ground truth data set, we obtain a histogram of region size measurement. The normalized frequency values of the histogram are used to determine the mass distribution.

(2) Texture. Texture refers to a description of the spatial variation within a contiguous group of pixels. There are small objects in forest and residential areas, such as trees, houses, roads and shadows. As a result, these regions show a great variety of color and brightness. They indicate a "high-contrast texture" area. On the other hand, crop fields, land, lakes and sea indicate a "fine texture" area. To determine the mass distribution of the texture feature, we measure the amount of edges per unit image area. The Roberts gradient is computed over image windows.

(3) Region Shapes. The region shape can be characterized by three features - FIT, ELONG and DIREC of the region. The minimum bounding rectangle (MBR) of a region is computed. The MBR is defined as the rectangle such that the ratio of the area of the region and the area of the enclosing rectangle with sides parallel to the coordinate axes is maximum, under rotations of 0 to 80 degrees. FIT is this maximum which measures the degree of matching of the region with rectangles. The elongatedness ELONG of a region is defined by ELONG = L/W, where L and W denote the lengths of the long and short sides of the MBR respectively. DIREC denotes the direction of the long side of the MBR. The mass distributions of these features are computed also by histograms of related measurements.

(4) Compactness. Another global shape feature is compactness of a region, which is defined by $4\pi area/perimeter^2$. Normally, objects with high compactness feature are candidates of man-made structures. The mass distribution is determined by histogram of feature measurements.

(5) Spectral Information. The spectral feature of a region is defined by the intensity mean vector. The mass distribution is similarly computed.

(6) Spatial Relationships. The adjacency graph describes the interrelationships of regions. A probability transition matrix is defined to provide feature measurements.

## 5. EXPERIMENTAL RESULTS

We have investigated multispectral images of crop fields at Clarke, Oregon. Eight classes of wheat, alfalfa, potatoes, corn, beans, apple, pasture and rangeland are selected from multispectral scanner data (Figure 1). Part of the selected data is used for training and a much larger part is used for testing. The accuracy of maximum likelihood classification performed by Thomas in 1982 (see [1]) is about 75 %. The contextual classification using a dynamic programming approach raised the classification accuracy to 80.5 % [1]. The contextual classification using the stochastic relaxation approach raised the classification accuracy to 80.8 % [2]. The contingency tables in Figures 2 give comparisons of these methods with the evidential reasoning approach which has a more than 2.5 % accuracy improvement.

## 6. CONCLUSION

In multispectral image understanding, it is very difficult to build an exact world model for the analysis of complex aerial photographs. There are uncertainty and incompleteness in information and knowledge. The evidential reasoning approach using the Dempster-Shafer theory is a powerful tool that proves to be useful in this application. Information from multiple sources are combined to reduce uncertainty and to obtain real world information.

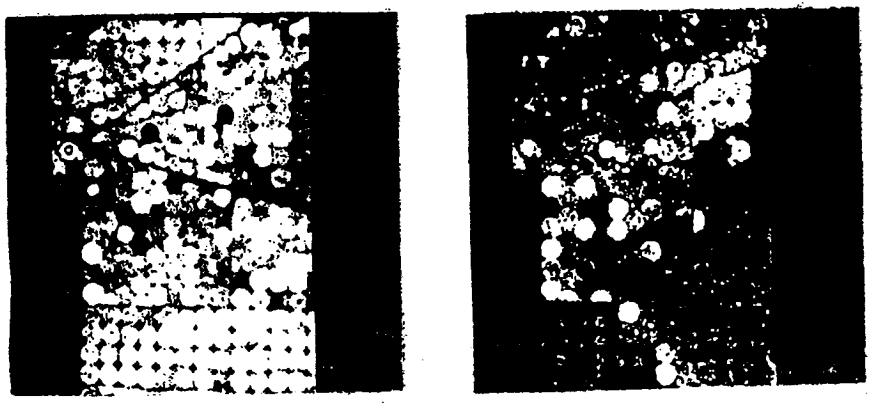

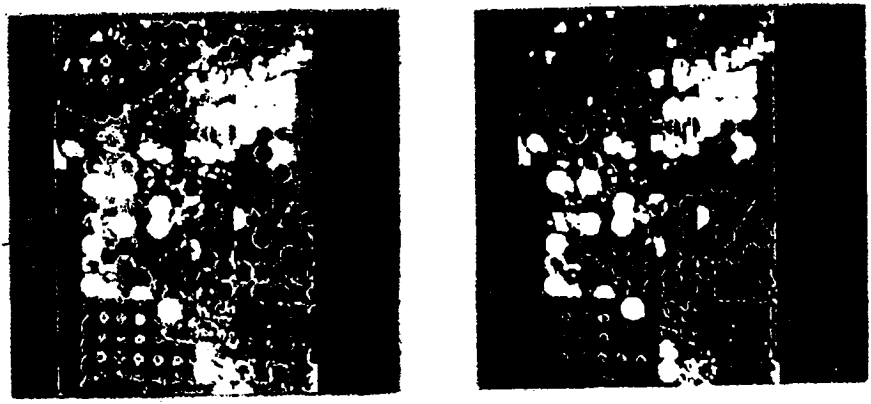

(a)    (b)

(c)    (d)

Figure 1. (a) First band of MSS scene of cropfield at Clarke, Oregon:1982. (b) Bayes Preclassification result (c) Contextual classification result by Stochastic Relaxation (d) Contextual classification using Dempster-Shafer approach.

Figure 2   Contingency tables for classification results
of test image 'Clark'. Scale factor of
the number of pixels $10^{**}1$.

COL = assigned categories    ROW = true categories

(A) Pixel independent Bayes classification result

| CLASS | WHT | ALF | POT | CRN | BNS | APL | PAS | RNG | TOTAL | ACC(%)* |
|---|---|---|---|---|---|---|---|---|---|---|
| WHT | 1017 | 47 | 30 | 5 | 4 | 0 | 10 | 75 | 1188 | 85.5% |
| ALF | 71 | 382 | 135 | 10 | 13 | 6 | 12 | 39 | 668 | 57.1% |
| POT | 40 | 32 | 522 | 5 | 19 | 0 | 2 | 32 | 652 | 84.6% |
| CRN | 1 | 5 | 1 | 65 | 2 | 0 | 0 | 4 | 78 | 83.3% |
| BNS | 0 | 1 | 0 | 1 | 1 | 0 | 0 | 0 | 3 | 0% |
| PAS | 0 | 0 | 0 | 0 | 0 | 0 | 9 | 2 | 11 | 81.1% |
| RNG | 15 | 12 | 14 | 2 | 4 | 1 | 9 | 335 | 392 | 85.4% |
| TOTAL | 1146 | 483 | 704 | 89 | 78 | 7 | 42 | 490 | 3040 | 77.5%** |

(B) Context classification result using a dynamic programming approach

| CLASS | WHT | ALF | POT | CRN | BNS | APL | PAS | RNG | TOTAL | ACC(%)* |
|---|---|---|---|---|---|---|---|---|---|---|
| WHT | 1073 | 26 | 26 | 1 | 11 | 0 | 10 | 601 | 1248 | 90.9% |
| ALF | 89 | 390 | 150 | 3 | 1 | 0 | 1 | 34 | 668 | 58.4% |
| POT | 58 | 29 | 534 | 2 | 6 | 0 | 0 | 23 | 652 | 81.9% |
| CRN | 1 | 5 | 1 | 68 | 0 | 0 | 0 | 4 | 79 | 86.1% |
| BNS | 1 | 6 | 2 | 1 | 36 | 0 | 0 | 3 | 49 | 73.5% |
| APL | 0 | 2 | 0 | 0 | 0 | 0 | 0 | 0 | 2 | 0% |
| PAS | 0 | 1 | 0 | 0 | 0 | 0 | 8 | 3 | 11 | 72.7% |
| RNG | 19 | 16 | 15 | 1 | 3 | 0 | 1 | 339 | 394 | 86.1% |
| TOTAL | 1681 | 605 | 777 | 82 | 58 | 0 | 23 | 1064 | 3040 | 80.5%** |

(C) Stochastic relaxation Context classification result

| CLASS | WHT | ALF | POT | CRN | BNS | APL | PAS | RNG | TOTAL | ACC(%)* |
|---|---|---|---|---|---|---|---|---|---|---|
| WHT | 1080 | 23 | 25 | 1 | 1 | 0 | 0 | 58 | 1118 | 90.9% |
| ALF | 91 | 378 | 155 | 1 | 0 | 0 | 0 | 41 | 666 | 56.8% |
| POT | 54 | 23 | 544 | 1 | 3 | 0 | 0 | 29 | 654 | 83.2% |
| CRN | 1 | 5 | 1 | 65 | 0 | 0 | 0 | 6 | 78 | 83.3% |
| BNS | 2 | 5 | 2 | 0 | 35 | 0 | 0 | 4 | 48 | 73.9% |
| APL | 1 | 2 | 0 | 0 | 0 | 0 | 0 | 0 | 3 | 0% |
| PAS | 0 | 1 | 0 | 0 | 0 | 0 | 7 | 3 | 11 | 63.7% |
| RNG | 17 | 11 | 14 | 1 | 0 | 0 | 1 | 349 | 392 | 89.1% |
| TOTAL | 1643 | 573 | 787 | 72 | 47 | 0 | 10 | 1158 | 3040 | 80.8%** |

(D) contextual classification using the Dempster-Shafer approach

| CLASS | WHT | ALF | POT | CRN | BNS | APL | PAS | RNG | TOTAL | ACC(%)* |
|---|---|---|---|---|---|---|---|---|---|---|
| WHT | 1096 | 11 | 14 | 0 | 1 | 0 | 0 | 40 | 1162 | 94.32% |
| ALF | 90 | 384 | 150 | 1 | 0 | 0 | 0 | 41 | 666 | 57.66% |
| POT | 38 | 19 | 569 | 0 | 2 | 0 | 0 | 27 | 655 | 86.87% |
| CRN | 1 | 5 | 1 | 71 | 0 | 0 | 0 | 6 | 80 | 84.52% |
| BNS | 2 | 5 | 2 | 0 | 42 | 0 | 0 | 4 | 55 | 76.36% |
| APL | 1 | 2 | 0 | 0 | 0 | 0 | 0 | 0 | 3 | 0% |
| PAS | 0 | 1 | 0 | 0 | 0 | 0 | 9 | 0 | 10 | 90% |
| RNG | 17 | 11 | 14 | 1 | 0 | 0 | 1 | 361 | 405 | 89.14% |
| TOTAL | 1245 | 438 | 750 | 73 | 44 | 0 | 10 | 479 | 3040 | 83.3% |

* Classification accuracy.
** Overall classification accuracy: ration of the number correctly classified pixels to the number of total classified pixels.

WHT — Wheat
ALF — Alfalfa
POT — Potatoes
CRN — Corn
RNS — Beans
APL — Apples
PAS — Pasture (irrigated)
RNG — Rangeland